\documentclass[10pt,twocolumn,letterpaper]{article}

\usepackage{cvpr}              
\definecolor{cvprblue}{rgb}{0.21,0.49,0.74}
\usepackage[pagebackref,breaklinks,colorlinks,allcolors=cvprblue]{hyperref}

\usepackage[accsupp]{axessibility}  

\usepackage{booktabs}
\usepackage{multirow}
\usepackage{siunitx}
\usepackage{relsize}
\usepackage{amssymb}
\usepackage{pifont}
\usepackage{bm}
\usepackage{subcaption}
\usepackage{mathtools}

\newcommand{\citefmt}[1]{\small(#1)}
\newcommand{\rankone}[1]{\textbf{#1}} 
\newcommand{\ranktwo}[1]{\underline{#1}}
\newcommand{\rankthree}[1]{\textit{#1}} 

\def\paperID{39734} 
\def\confName{CVPR}
\def\confYear{2026}

\title{Long-SCOPE: Fully Sparse Long-Range Cooperative 3D Perception}

\author{Jiahao Wang\textsuperscript{\rm 1}, 
Zikun Xu\textsuperscript{\rm 1}, 
Yuner Zhang\textsuperscript{\rm 2}, 
Zhongwei Jiang\textsuperscript{\rm 3}, 
Chenyang Lu\textsuperscript{\rm 1}, 
Shuocheng Yang\textsuperscript{\rm 1}\\
Yuxuan Wang\textsuperscript{\rm 1}, 
Jiaru Zhong\textsuperscript{\rm 4}, 
Chuang Zhang\textsuperscript{\rm 1}, 
Shaobing Xu\textsuperscript{\rm 1}\thanks{Corresponding authors}
\addtocounter{footnote}{-1}, 
Jianqiang Wang\textsuperscript{\rm 1}\footnotemark\\
\textsuperscript{\rm 1}Tsinghua University, 
\textsuperscript{\rm 2}University of Pennsylvania\\
\textsuperscript{\rm 3}Nanyang Technological University, 
\textsuperscript{\rm 4}The Hong Kong Polytechnic University\\
{\tt\small wjh22@mails.tsinghua.edu.cn, \{shaobxu, wjqlws\}@tsinghua.edu.cn}
}

\begin{document}
\maketitle
\begin{abstract}
Cooperative 3D perception via Vehicle-to-Everything communication is a promising paradigm for enhancing autonomous driving, offering extended sensing horizons and occlusion resolution.
However, the practical deployment of existing methods is hindered at long distances by two critical bottlenecks:
the quadratic computational scaling of dense BEV representations and the fragility of feature association mechanisms under significant observation and alignment errors.
To overcome these limitations, we introduce Long-SCOPE, a fully sparse framework designed for robust long-distance cooperative 3D perception.
Our method features two novel components: a Geometry-guided Query Generation module to accurately detect small, distant objects,
and a learnable Context-Aware Association module that robustly matches cooperative queries despite severe positional noise.
Experiments on the V2X-Seq and Griffin datasets validate that Long-SCOPE achieves state-of-the-art performance, particularly in challenging 100--150~m long-range settings, while maintaining highly competitive computation and communication costs.
\end{abstract}    
\section{Introduction}
\label{sec:introduction}

Despite substantial advancements in autonomous driving technologies~\cite{xuMergeOccBridgeDomain2025,yangRINOAccurateRobust2025,hanEquiRO4DMmWave2026}, current single-vehicle systems remain fundamentally constrained by limited sensor fields-of-view, performance degradation at long ranges, and severe occlusions.
To overcome these individual limitations, cooperative perception has emerged as a key paradigm.
By enabling information exchange between multiple agents such as vehicles (V2V), infrastructure (V2I), and drones (V2D), it promises to compensate for the deficiencies of single-vehicle systems, particularly by resolving occlusions and extending the effective perception horizon.

\begin{figure}[t]
    \centering
    \includegraphics[width=\linewidth]{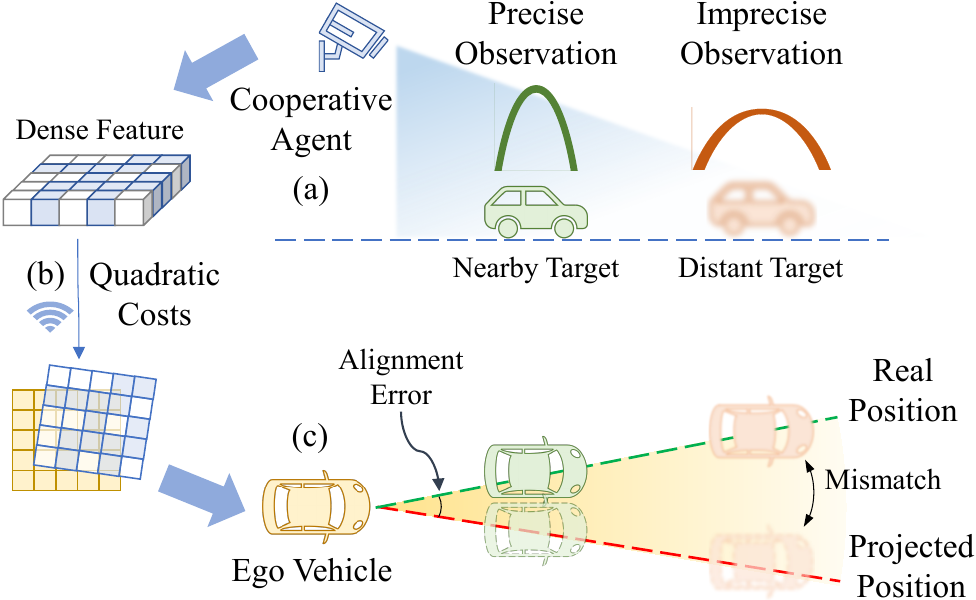}
    \caption{Core challenges in long-range cooperation: (a) observation imprecision for distant targets, (b) quadratic scaling costs of dense features, and (c) positional mismatch from alignment errors.}
    \label{fig:1_motivation}
    \vspace{-0.3em}
\end{figure}

However, a fundamental conflict between performance and cost limits the practical deployment of this vision.
Effective cooperation necessitates a high density of participating agents, yet prevailing methods often depend on expensive LiDAR sensors, failing to leverage already widespread, low-cost roadside cameras.
Moreover, the associated communication and computational demands impose heavy hardware infrastructure costs.
Therefore, a critical bottleneck to real-world deployment is maximizing long-range perception performance using low-cost visual sensors within strictly limited communication and computational budgets.
This introduces two primary challenges:

First is the choice of information structure for transmission.
As shown in~\cref{fig:1_motivation}(b), mainstream cooperative perception methods rely on dense BEV features~\cite{wangV2VNetVehicletovehicleCommunication2020,liLearningDistilledCollaboration2021,huCollaborationHelpsCamera2023, luRobustCollaborative3D2023},
for which both computation and communication costs scale quadratically with the perception range.
Although recent methods transmitting sparse feature structures have enabled communication costs to scale linearly with the number of targets in the scene, their frameworks still depend on dense BEV backbones~\cite{liuSparseCommEfficientSparse2025,wangIFTRInstanceLevelFusion2025, yuanSparseAlignFullySparse2025, wangCoopDETRUnifiedCooperative2025,zhongCoopTrackExploringEndtoEnd2025}, thereby inheriting the same computational scaling limitations.

\begin{figure}[t]
    \centering
    \includegraphics[width=\linewidth]{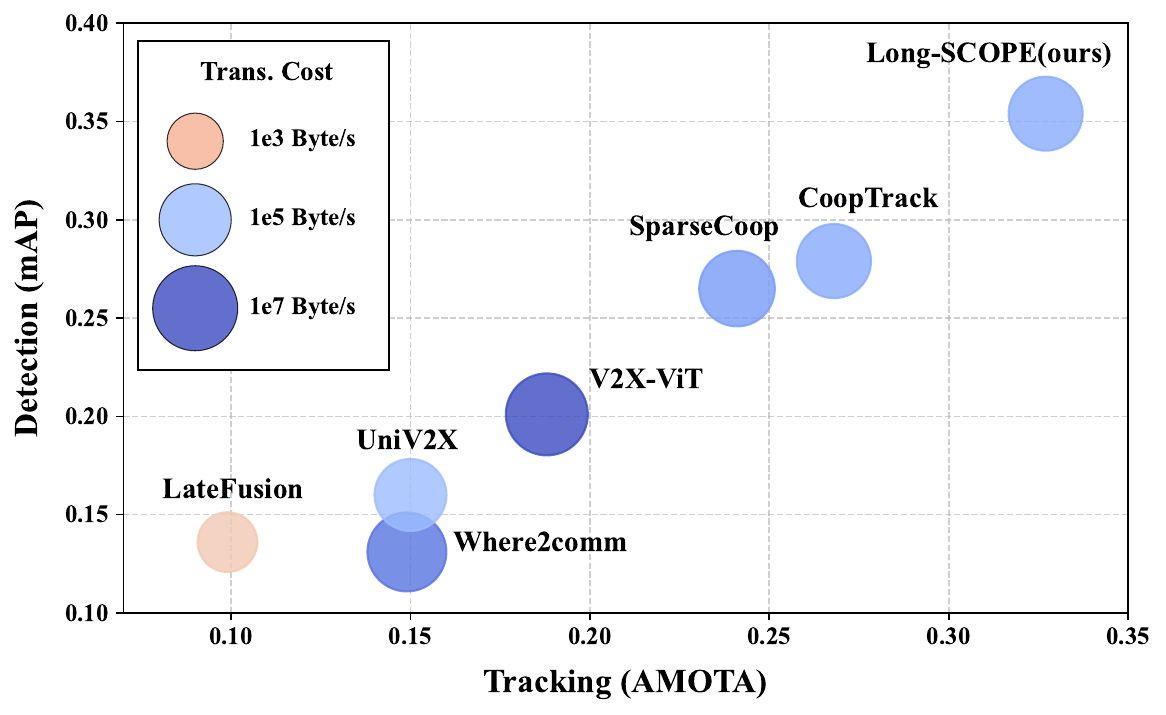}
    \caption{Performance comparison on \textit{Griffin-25m} dataset.
    The X-axis and Y-axis represent perception metrics,
    while the bubble size and color encode the transmission cost on a logarithmic scale.
    }
    \label{fig:1_performance}
\end{figure}

Second is the difficulty of accurately fusing cooperative information after transmission.
As shown in~\cref{fig:1_motivation}(a), for distant targets, the information extracted and sent by each agent inherently carries greater observation error.
Simultaneously, calibration errors and localization noise between agents create alignment errors, leading to severe positional deviations for distant targets during coordinate projection (\cref{fig:1_motivation}(c)).
Most existing cooperative fusion methods rely on fixed distance thresholds, which are highly fragile under the influence of these two factors, leading to incorrect matches and increased ambiguity.

To address these problems, we propose \textbf{Long-SCOPE}, a \textbf{Long}-Range \textbf{S}parse \textbf{CO}operative \textbf{PE}rception framework designed to push cooperative perception toward practical, long-range application.
Our method completely abandons BEV features, adopting a fully sparse architecture that extracts object queries directly from image features.
The main contributions are summarized as follows:
\begin{itemize}
    \item To reduce observation error in distant targets, we introduce a \textbf{Geometry-guided Query Generation} (GQG) module.
    This component utilizes 2D and position priors to dynamically generate accurately located 3D queries at the single-agent level, significantly enhancing initial detection quality.
    \item To overcome fusion failures from alignment errors, we design a novel \textbf{Context-Aware Association} (CAA) module.
    It leverages both spatial context and semantic similarity of queries to achieve robust matching, resolving cooperative-level ambiguity at long distances and demonstrating superior resilience to positional noise.
    \item We are the first to establish comprehensive long-range visual perception benchmarks on the \textit{V2X-Seq}~\cite{yuV2XseqLargescaleSequential2023} and \textit{Griffin-25m}~\cite{wangGriffinAerialGroundCooperative2026} datasets.
    We rigorously evaluate different cooperative paradigms and demonstrate that Long-SCOPE achieves state-of-the-art (SOTA) performance, setting a new standard for both accuracy and efficiency, as highlighted in~\cref{fig:1_performance}.
\end{itemize}

\section{Related Work}
\label{sec:related_work}

Cooperative perception strategies are broadly categorized into data-level (early fusion)~\cite{valienteControllingSteeringAngle2019,chenCooperCooperativePerception2019,arnoldCooperativePerception3D2022}, result-level (late fusion)~\cite{chiuProbabilistic3DMultiObject2024}, and feature-level (intermediate fusion).
Due to the prohibitive communication costs of early fusion and the information loss from early quantization in late fusion, feature-level fusion has become the predominant trade-off.
However, mainstream feature-level methods rely on dense Bird's-Eye-View (BEV) representations~\cite{wangV2VNetVehicletovehicleCommunication2020,luRobustCollaborative3D2023,chenCoopDiffDiffusionGuidedApproach2026,chenCATNetCollaborativeAlignment2026}, imposing quadratic computational and communication scaling relative to the perception range.
While compression~\cite{huCommunicationefficientCollaborativePerception2024,weiInfoComKilobyteScaleCommunicationEfficient2026}
and salient region selection~\cite{huWhere2commCommunicationefficientCollaborative2022,yuanGeneratingEvidentialBEV2023}
reduce bandwidth, the fundamental quadratic computational bottleneck remains.

To address this, an efficient paradigm~\cite{chenTransIFFInstancelevelFeature2023, fanQUESTQueryStream2024,zhongLeveragingTemporalContexts2024} using sparse, object-level queries has emerged,
achieving communication costs that scale linearly with the number of targets rather than the perception range.
However, many of these "sparse-communication" frameworks still depend on dense BEV backbones for query generation~\cite{liuSparseCommEfficientSparse2025,wangIFTRInstanceLevelFusion2025, yuanSparseAlignFullySparse2025, wangCoopDETRUnifiedCooperative2025,zhongCoopTrackExploringEndtoEnd2025},
reintroducing the quadratic computational bottleneck.
Recently, SparseCoop~\cite{wangSparseCoopCooperativePerception2026} introduced the first fully sparse architecture, which serves as our baseline.

\begin{figure}[b]
    \centering
    \includegraphics[width=0.6\linewidth]{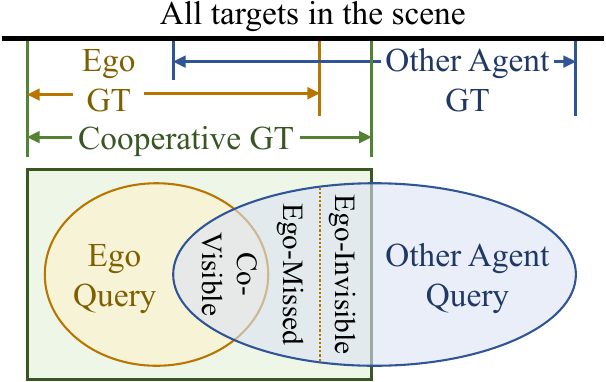}
    \caption{
    Task Definition of Cooperative Perception.
    Co-Visible, Ego-Missed, and Ego-Invisible cases should be correctly handled.
    }
    \label{fig:3_0_definition}
\end{figure}

\begin{figure*}[t]
    \centering
    \includegraphics[width=0.98\textwidth]{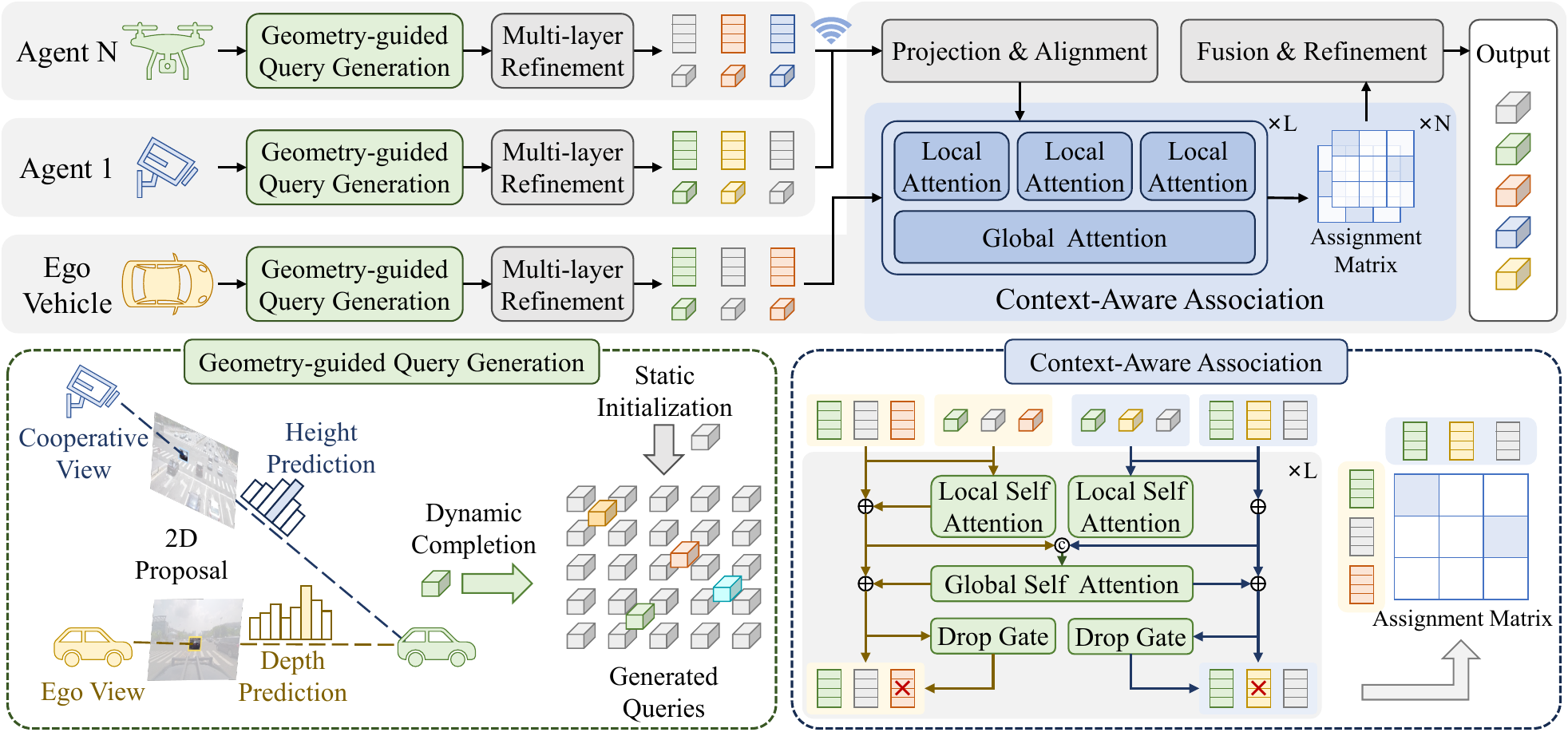}
    \caption{Overview of Long-SCOPE framework, highlighting our novel components: the Geometry-guided Query Generation module for dynamically proposing object queries and the Context-Aware Association module for robust multi-agent query matching.}
    \label{fig:3_overview}
\end{figure*}

A critical remaining challenge in these sparse frameworks is the robust association of queries, particularly in the presence of long-range positional mismatches.
Early methods~\cite{chenTransIFFInstancelevelFeature2023,liuSparseCommEfficientSparse2025} naively scattered cooperative queries into the ego BEV grid, largely ignoring observation errors.
Subsequent approaches employed explicit pairwise matching based on
relative distance~\cite{dingPointClusterCompact2024,zhongLeveragingTemporalContexts2024,yuEndtoEndAutonomousDriving2025},
feature similarity~\cite{wangCoopDETRUnifiedCooperative2025,wangSparseCoopCooperativePerception2026},
or position embeddings~\cite{yuanSparseAlignFullySparse2025,zhongCoopTrackExploringEndtoEnd2025}.
To mitigate the fragility of fixed distance thresholds~\cite{dingPointClusterCompact2024,wangCoopDETRUnifiedCooperative2025},
recent works~\cite{zhongLeveragingTemporalContexts2024,yuEndtoEndAutonomousDriving2025,yuanSparseAlignFullySparse2025,wangSparseCoopCooperativePerception2026} employ bipartite matching via the Hungarian algorithm~\cite{kuhnHungarianMethodAssignment1955} for global optimization.
However, a fundamental limitation persists: the handling of non-corresponding queries (e.g., ego-missed or invisible targets in~\cref{fig:3_0_definition}). 
Even globally optimized methods typically resort to a final, non-adaptive distance threshold to reject unmatched queries.
Under the significant positional noise inherent to long distances, this rigid heuristic inadvertently filters out valid cooperative queries.
Long-SCOPE directly addresses this precise vulnerability of previous methods.
\section{Method}
\label{sec:method}

\subsection{Preliminaries: Task Definition}
\label{sec:preliminaries}

We adopt the standard task formulation from prominent cooperative perception benchmarks~\cite{yuDAIRV2XLargescaleDataset2022,yuV2XseqLargescaleSequential2023,wangGriffinAerialGroundCooperative2026}.
As shown in~\cref{fig:3_0_definition} (Top), the cooperative perception ground truth $\mathit{GT}$ is achieved by unifying annotations from the ego-agent $\mathit{GT}_{\text{ego}}$ and cooperative agents $\mathit{GT}_{\text{co}}$,
which are transformed into the ego-agent's coordinate system, denoted as $\widetilde{\mathit{GT}}_{\text{co}}$.
The resulting set is then filtered to include only objects within an ego-centered Region of Interest (ROI), $R_{\text{ego}}$.
The final set is defined as:
$$ \mathit{GT} = \{ o \mid o \in \mathit{GT}_{\text{ego}} \cup \widetilde{\mathit{GT}}_{\text{co}}, \; c(o) \in R_{\text{ego}} \} $$
where $o$ is a ground truth object and $c(o)$ is its geometric center.
As shown in~\cref{fig:3_0_definition} (Bottom),
achieving this $\mathit{GT}$ requires not only associating co-visible queries under positional noise,
but also correctly integrating ego-missed and ego-invisible queries to enrich the scene,
all while avoiding the creation of duplicate detections.

\subsection{Pipeline Overview}
\label{sec:overview}

Long-SCOPE is a fully sparse, query-centric framework that circumvents the quadratic scaling costs of traditional dense BEV feature maps.
As illustrated in~\cref{fig:3_overview}, our pipeline proceeds as follows:
First, at the single-agent level, each agent initiates a set of object queries.
This set is a consolidated collection of both static, fixed anchors and dynamic queries generated by our novel Geometry-guided Query Generation (GQG) module.
These combined queries are then processed by a multi-layer transformer decoder~\cite{linSparse4DV3Advancing2023} to refine their semantic features and state vectors.
The queries from cooperative agents are then projected and aligned into the ego-vehicle's coordinate system to compensate for spatio-temporal differences.
Next, all queries—both local and aligned cooperative—are fed into our Context-Aware Association (CAA) module, which robustly matches corresponding queries and identifies unique (e.g., ego-invisible) objects.
Finally, the resulting set of matched, unmatched, and local queries is processed by a fusion and refinement network to produce the comprehensive 3D detection output.

Our architecture leverages the foundational sparse paradigm of SparseCoop~\cite{wangSparseCoopCooperativePerception2026}, adopting its core modules for multi-layer refinement, spatio-temporal alignment, and final fusion.
However, this baseline is fragile under long-range conditions, a limitation rooted in its core association logic.
To maintain performance, it relies on a simple heuristic: associating only cooperative queries within a 30m or even 15m radius of the ego-vehicle, while naively forwarding all queries beyond this range without any association.
This simplistic approach suffers from duplicate detections and fails to resolve the significant observation and alignment errors inherent in long-distance perception.

To overcome these limitations, Long-SCOPE introduces two novel components that replace the baseline's most vulnerable modules.
First, to address observation errors, the GQG module (\cref{sec:geometry-guided-query-generation}) augments static anchors with dynamically generated queries informed by geometric priors, significantly improving initial detection quality for distant targets.
Second, to resolve projection errors and fragile matching, the CAA module (\cref{sec:context-aware-association}) replaces rule-based association with a learnable, attention-based mechanism.
By leveraging both spatial context and semantic similarity, it achieves robust query association well beyond the 30m limit, forming the core of our long-range fusion capability.

\subsection{Geometry-guided Query Generation}
\label{sec:geometry-guided-query-generation}

The conventional approach of initializing object queries from a fixed set of static anchors~\cite{wangDETR3D3DObject2022,wangSparseCoopCooperativePerception2026} is suboptimal for long-range perception.
Distant objects are small and sparsely distributed, meaning most static queries will be initialized far from any valid target, compromising perception performance.

To address this, we introduce the Geometry-guided Query Generation module,
which aims to dynamically generate a set of high-quality, data-driven queries by leveraging 2D image priors.
Similar to Far3D~\cite{jiangFar3DExpandingHorizon2024}, we first employ lightweight 2D detection and monocular depth estimation heads on the shared image features.
For a target Q in the scene, these heads provide a coarse 2D proposal centered at pixel $P_{\text{img}} = [u, v, 1]^T$ and an initial depth estimate $\hat{z}_{Q_{\text{cam}}}$.

\begin{figure}[t]
    \centering
    \includegraphics[width=\linewidth]{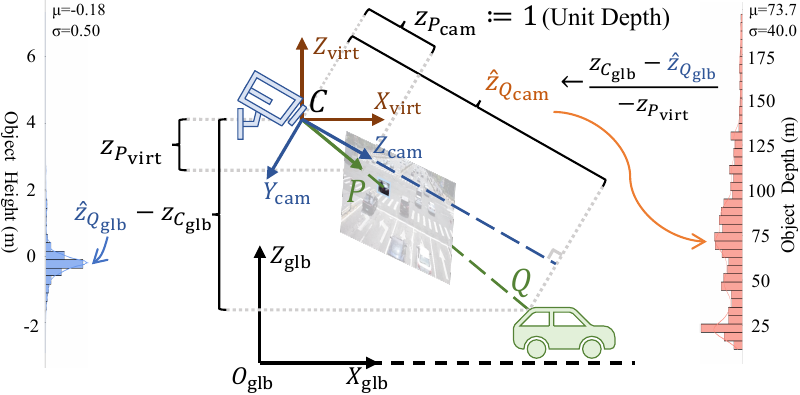}
    \caption{Depth derivation for high-vantage agents.
    We predict the stable global height $\hat{z}_{Q_{\text{glb}}}$ (left histogram) instead of the dispersed depth $\hat{z}_{Q_{\text{cam}}}$ (right histogram) and utilize geometric priors.
    }
    \label{fig:3_1_transform}
\end{figure}

However, direct depth regression is an ill-posed problem.
Inspired by BEVHeight~\cite{yangBEVHeightRobustFramework2023,yangBEVHeightRobustVisual2025} and DHD~\cite{wangDronesHelpDrones2024},
for agents with a high vantage point (e.g., roadside units, drones), we can simplify this task by instead predicting the object's global height, $\hat{z}_{Q_{\text{glb}}}$.
As illustrated by the statistics in~\cref{fig:3_1_transform}, the object height distribution in typical driving scenes is highly concentrated,
whereas the depth distribution is extremely dispersed.
This makes global height a significantly more stable and well-posed target for optimization.

Consequently, for high-vantage agents, we re-frame the depth estimation problem as a geometric derivation from this more stable height prediction.
This derivation requires three coordinate frames:
(1) the global frame $\mathcal{F}_{\text{glb}}$ (e.g., East-North-Up);
(2) the camera frame $\mathcal{F}_{\text{cam}}$, with its origin $C$ at position $C_{\text{glb}}$ in $\mathcal{F}_{\text{glb}}$, where the projection matrix $T_{\text{cam2glb}} \in \mathbb{R}^{4 \times 4}$ is known from calibration and localization data;
and (3) a "virtual" frame $\mathcal{F}_{\text{virt}}$, also centered at $C$ but with axes parallel to $\mathcal{F}_{\text{glb}}$, sharing the rotation $R_{\text{cam2virt}} \in \mathbb{R}^{3 \times 3}$ from $T_{\text{cam2glb}}[:3, :3]$.

Our 2D-to-3D lifting for high-vantage agents, diagrammed in~\cref{fig:3_1_transform}, begins with a network head predicting the target's global height, $\hat{z}_{Q_{\text{glb}}}$, for a 2D proposal at pixel $P_{\text{img}}$.
First, we compute the target's height in the virtual frame, $\hat{z}_{Q_{\text{virt}}} = \hat{z}_{Q_{\text{glb}}} - z_{C_{\text{glb}}}$, by subtracting the camera's known global height $z_{C_{\text{glb}}}$.
Next, we find the corresponding ray's virtual height by un-projecting the pixel $P_{\text{img}}$ to a 3D ray in the camera frame, $P_{\text{cam}} = K_{\text{cam}}^{-1} \cdot P_{\text{img}}$, assuming unit depth ($z_{P_{\text{cam}}} \coloneqq 1$).
This ray is transformed into the virtual frame via $P_{\text{virt}} = R_{\text{cam2virt}} \cdot P_{\text{cam}}$, and we extract its $z$-component, $z_{P_{\text{virt}}}$.
As shown in~\cref{fig:3_1_transform}, the geometry forms two similar triangles, allowing us to equate the ratio of the target's camera depth $\hat{z}_{Q_{\text{cam}}}$ to the unit depth (1) with the ratio of the target's virtual height $\hat{z}_{Q_{\text{virt}}}$ to the ray's virtual height $z_{P_{\text{virt}}}$.
We can therefore solve for the unknown depth $\hat{z}_{Q_{\text{cam}}}$ directly:
\begin{equation}
    \hat{z}_{Q_{\text{cam}}} = \frac{\hat{z}_{Q_{\text{virt}}}}{z_{P_{\text{virt}}}} = \frac{\hat{z}_{Q_{\text{glb}}} - z_{C_{\text{glb}}}}{(T_{\text{cam2glb}}[:3, :3] \cdot K_{\text{cam}}^{-1} \cdot P_{\text{img}})_z}
    \label{eq:depth-derivation}
\end{equation}
This derivation provides a stable depth estimate. For high-angle views, $z_{P_{\text{virt}}}$ is significantly non-zero, avoiding numerical issues.

Conversely, for ground-level agents (e.g., vehicles), the camera view is nearly parallel to the ground plane.
For pixels near the horizon, $z_{P_{\text{virt}}} \approx 0$, making the division in~\cref{eq:depth-derivation} numerically unstable.
Therefore, for these agents, we revert to the standard approach and use a lightweight head $\mathcal{D}$ to directly regress the depth $\hat{z}_{Q_{\text{cam}}} = \mathcal{D}(f_{\text{img}}(P_{\text{img}}))$ using the image features $f_{\text{img}}(P_{\text{img}})$.

Regardless of which estimation strategy is employed, we obtain a 3D position $P_{\text{cam}}$ in the camera frame $\mathcal{F}_{\text{cam}}$ by un-projecting the 2D proposal $P_{\text{img}}$ with its estimated depth $\hat{z}_{Q_{\text{cam}}}$ as $P_{\text{cam}} = \hat{z}_{Q_{\text{cam}}} \cdot (K_{\text{cam}}^{-1} P_{\text{img}})$.
This point is then transformed into the agent's local coordinate system using known camera extrinsics.
This 3D position, along with the corresponding image feature extracted via MaxPooling, are used to initialize a new dynamic query.
These dynamically generated queries are then merged with the set of static, fixed-initialized queries and fed collectively into the multi-layer 3D decoder,
significantly improving its ability to find and refine distant targets.

\subsection{Context-Aware Association}
\label{sec:context-aware-association}

The core bottleneck in long-range cooperative fusion lies in the robust association of sparse queries.
As demonstrated by the baseline limitations in~\cref{sec:overview},
simple heuristics such as distance-based cutoffs are highly sensitive to the significant observation and projection noise inherent to distant targets.
This fragility results in duplicate detections from unassociated queries and a failure to capitalize on the extended field-of-view that cooperation aims to deliver.
To overcome this, we introduce the Context-Aware Association module, which replaces rigid heuristics with a learnable, context-driven matching mechanism.

To ensure robustness against long-range ambiguities, its architecture is governed by four core design principles:
\begin{itemize}
    \item \textbf{Unambiguous Matching:}
    A single target should generate at most one query per agent.
    Therefore, valid associations should strictly enforce injective (one-to-one) correspondences, preventing a single query from being matched multiple times.
    \item \textbf{Asymmetric Visibility:}
    The formulation must natively support unmatchable queries arising from ego-missed or ego-invisible targets, seamlessly preserving them for the final perception output.
    \item \textbf{Spatial Consistency:}
    While the absolute coordinates of distant targets are frequently corrupted by alignment errors, their local spatial context remains a highly stable signature.
    A successful match must demonstrate consistency not only in semantic features but also within its local neighborhood topology.
    \item \textbf{Scalability:}
    The association framework must generalize efficiently to a dynamic number of $N$ cooperative agents, extending beyond simple pairwise matching.
\end{itemize}

From this perspective, existing rule-based approaches with fixed distance thresholds~\cite{dingPointClusterCompact2024,wangCoopDETRUnifiedCooperative2025} satisfy only the Asymmetric Visibility requirement.
While set-based bipartite matching~\cite{zhongLeveragingTemporalContexts2024,yuEndtoEndAutonomousDriving2025,wangSparseCoopCooperativePerception2026} resolves Unambiguous Matching, it still ignores Spatial Consistency and is generally confined to pairwise fusion without Scalability.

To comprehensively satisfy these principles, our CAA module employs a multi-layer Transformer~\cite{vaswaniAttentionAllYou2017} architecture inspired by recent advancements in robust feature matching~\cite{sarlinSuperGlueLearningFeature2020,lindenbergerLightGlueLocalFeature2023}.
Unlike standard cross-attention mechanisms restricted to two sets, we further adopt an extensible Structure-from-Motion (SfM) approach~\cite{wangVGGTVisualGeometry2025,wangP3PermutationEquivariantVisual2025}:
we concatenate the queries from all $N$ agents and utilize global self-attention.
While this incurs a quadratic computational cost relative to the query count, the total number of high-confidence transmitted queries is tied to the physical targets in the scene, generally remaining below 100 and ensuring high efficiency.

\noindent\textbf{Contextual Feature Refinement.}
As shown in~\cref{fig:3_overview}, the CAA module refines query descriptors iteratively over $L$ layers.
Each layer consists of two sequential operations:
First, to encode \textit{Spatial Consistency}, an intra-agent (local) self-attention block operates independently on each agent's query set.
This block utilizes positional encodings derived from the queries' state vectors,
allowing each query to aggregate topological information from its spatial neighbors.
This enriches the descriptor with a stable relative context of its surrounding targets.
Second, an inter-agent (global) self-attention block operates across the concatenated pool of all $N$ agents' queries.
Crucially, positional encoding is disabled here to achieve invariance to projection noise.
This unconstrained global attention enables queries representing the same physical object—now augmented with robust local topology—to implicitly identify correspondences and mutually align their semantic representations.

\noindent\textbf{Optimal Transport Matching.}
After $L$ refinement layers, the module produces highly contextualized query descriptors.
To extract discrete associations, we decouple the global multi-way matching into parallelized pairwise assignment heads:
for each cooperative agent $i$, we compute an affinity matrix $S$ between its refined queries $\mathcal{Q}_{coop,i}$ and the ego queries $\mathcal{Q}_{ego}$ using feature inner products.
Following SuperGlue~\cite{sarlinSuperGlueLearningFeature2020}, we apply Sinkhorn normalization~\cite{sinkhornConcerningNonnegativeMatrices1967} to $S$ to derive a doubly stochastic partial assignment matrix $P(u,x)$, representing the matching probability between cooperative query $u$ and ego query $x$.
To yield the final assignment matrix, we scale these probabilities by per-descriptor matchability scores $(\sigma^u, \sigma^x)$ as formulated in LightGlue~\cite{lindenbergerLightGlueLocalFeature2023}.
The resulting matches are filtered via a mutual nearest-neighbor check and a fixed confidence threshold.
Matched queries proceed to the fusion and refinement stage, while the unmatched cooperative queries, representing uniquely observed ego-missed or ego-invisible targets, are directly appended to the final output.
\section{Experiments}
\label{sec:experiments}

\begin{table*}[t]
    \centering
    \caption{
    Performance comparison on \textit{Griffin-25m}.
    \rankone{Bold}, \ranktwo{underline}, and \rankthree{italic} represent the top three among feature-level fusion methods.
    }
    \label{tab:griffin-25m}
    
    \setlength{\tabcolsep}{5pt} 

    \begin{tabular}{lccccccccc}
    \toprule
    \multirow{2}{*}[-0.5ex]{\textbf{Method}} & \multicolumn{2}{c}{\textbf{0--100 m Overall}} & \multicolumn{2}{c}{\textbf{0--50 m}} & \multicolumn{2}{c}{\textbf{50--100 m}} & \multirow{2}{*}[-0.5ex]{\textbf{BPS} $\downarrow$} & \multirow{2}{*}[-0.5ex]{\textbf{FPS} $\uparrow$} \\
    \cmidrule(lr){2-3} \cmidrule(lr){4-5} \cmidrule(lr){6-7}
     & \textbf{AP} $\uparrow$ & \textbf{AMOTA} $\uparrow$ & \textbf{AP} $\uparrow$ & \textbf{AMOTA} $\uparrow$ & \textbf{AP} $\uparrow$ & \textbf{AMOTA} $\uparrow$ & & \\
    \midrule

    Late Fusion Baseline & 0.136 & 0.099 & 0.357 & 0.325 & 0.004 & 0.000 & \num{2.55e3} & 6.13 \\
    \midrule

    Where2Comm~\cite{huWhere2commCommunicationefficientCollaborative2022} \citefmt{NIPS 2022} & 0.131 & 0.149 & 0.384 & 0.438 & 0.000 & 0.000 & \num{1.05e6} & 7.34 \\
    V2X-ViT~\cite{xuV2XViTVehicletoeverythingCooperative2022} \citefmt{ECCV 2022} & 0.201 & 0.188 & \rankthree{0.494} & \rankthree{0.527} & 0.011 & 0.000 & \num{3.21e6} & \rankthree{7.50} \\
    UniV2X~\cite{yuEndtoEndAutonomousDriving2025} \citefmt{AAAI 2025} & 0.160 & 0.150 & 0.417 & 0.428 & 0.007 & 0.000 & \rankone{$\mathbf{9.84 \times 10^4}$} & 6.51 \\
    CoopTrack~\cite{zhongCoopTrackExploringEndtoEnd2025} \citefmt{ICCV 2025} & \ranktwo{0.279} & \ranktwo{0.268} & \ranktwo{0.560} & \rankone{0.617} & \rankthree{0.054} & 0.000 & \rankthree{$\mathit{2.00 \times 10^5}$} & 6.03 \\
    SparseCoop~\cite{wangSparseCoopCooperativePerception2026} \citefmt{AAAI 2026} & \rankthree{0.265} & \rankthree{0.241} & \rankthree{0.494} & 0.478 & \ranktwo{0.088} & \ranktwo{0.016} & \num{3.31e5} & \rankone{8.87} \\
    \textbf{Long-SCOPE \citefmt{Ours}} & \rankone{0.354} & \rankone{0.327} & \rankone{0.597} & \ranktwo{0.566} & \rankone{0.151} & \rankone{0.112} & \ranktwo{\num{1.90e+05}} & \ranktwo{7.68} \\
    \midrule
    Long-SCOPE w/o GQG & 0.284 & 0.232 & 0.523 & 0.480 & 0.083 & 0.011 & \num{3.31e5} & 8.05 \\
    Long-SCOPE w/o CAA & 0.329 & 0.294 & 0.542 & 0.495 & 0.155 & 0.131 & \num{1.90e+05} & 8.10 \\
    \bottomrule
    \end{tabular}
\end{table*}

\begin{table*}[t]
    \centering
    \caption{
    Performance comparison on \textit{V2X-Seq}.
    \rankone{Bold}, \ranktwo{underline}, and \rankthree{italic} represent the top three among feature-level fusion methods.
    }
    \label{tab:v2x-seq}
    
    \setlength{\tabcolsep}{5pt} 

    \begin{tabular}{lcccccccc}
    \toprule
    \multirow{2}{*}[-0.5ex]{\textbf{Method}} & \multicolumn{2}{c}{\textbf{0--150 m Overall}} & \multicolumn{2}{c}{\textbf{0--50 m}} & \multicolumn{2}{c}{\textbf{50--100 m}} & \multicolumn{2}{c}{\textbf{100--150 m}} \\
    \cmidrule(lr){2-3} \cmidrule(lr){4-5} \cmidrule(lr){6-7} \cmidrule(lr){8-9}
     & \textbf{AP} $\uparrow$ & \textbf{AMOTA} $\uparrow$ & \textbf{AP} $\uparrow$ & \textbf{AMOTA} $\uparrow$ & \textbf{AP} $\uparrow$ & \textbf{AMOTA} $\uparrow$ & \textbf{AP} $\uparrow$ & \textbf{AMOTA} $\uparrow$ \\
    \midrule

    Late Fusion Baseline & 0.224 & 0.271 & 0.318 & 0.375 & 0.178 & 0.185 & 0.050 & 0.032 \\
    \midrule

    Where2Comm~\cite{huWhere2commCommunicationefficientCollaborative2022} \citefmt{NIPS 2022} & 0.059 & 0.004 & 0.141 & 0.080 & 0.026 & 0.000 & 0.000 & 0.000 \\
    V2X-ViT~\cite{xuV2XViTVehicletoeverythingCooperative2022} \citefmt{ECCV 2022} & \rankthree{0.289} & \ranktwo{0.345} & \rankthree{0.421} & \rankthree{0.509} & \ranktwo{0.231} & \ranktwo{0.251} & \rankthree{0.054} & \ranktwo{0.027} \\
    UniV2X~\cite{yuEndtoEndAutonomousDriving2025} \citefmt{AAAI 2025} & 0.166 & 0.019 & 0.255 & 0.101 & 0.123 & 0.000 & 0.032 & 0.000 \\
    CoopTrack~\cite{zhongCoopTrackExploringEndtoEnd2025} \citefmt{ICCV 2025} & 0.232 & 0.171 & 0.327 & 0.255 & 0.192 & 0.088 & \ranktwo{0.059} & \rankthree{0.006} \\
    SparseCoop~\cite{wangSparseCoopCooperativePerception2026} \citefmt{AAAI 2026} & \ranktwo{0.334} & \rankthree{0.328} & \ranktwo{0.515} & \ranktwo{0.534} & \rankthree{0.229} & \rankthree{0.232} & 0.031 & 0.000 \\
    \textbf{Long-SCOPE \citefmt{Ours}} & \rankone{0.399} & \rankone{0.444} & \rankone{0.532} & \rankone{0.576} & \rankone{0.339} & \rankone{0.361} & \rankone{0.113} & \rankone{0.059} \\
    \midrule
    Long-SCOPE w/o GQG & 0.388 & 0.416 & 0.524 & 0.559 & 0.321 & 0.341 & 0.113 & 0.046 \\
    Long-SCOPE w/o CAA & 0.352 & 0.382 & 0.530 & 0.588 & 0.243 & 0.280 & 0.041 & 0.005 \\
    \bottomrule
    \end{tabular}
\end{table*}

\subsection{Datasets and Metrics}
\label{subsec:datasets-and-metrics}

\noindent\textbf{V2X-Seq.}
This is a large-scale, sequential dataset designed for V2I cooperative 3D object detection and tracking~\cite{yuV2XseqLargescaleSequential2023}.
It features sensor data from both an ego-vehicle and a connected RSU, capturing complex urban traffic scenarios with significant occlusions.
We adhere to the official data splits and protocol of the CVPR 2025 challenge~\cite{haoResearchChallengesProgress2025}.
Critically, to rigorously benchmark the long-range performance central to our work, we expand the evaluation scope from the standard 50 meters to 150 meters.

\noindent\textbf{Griffin.}
This is a pioneering simulated dataset for aerial-ground cooperative perception~\cite{wangGriffinAerialGroundCooperative2026}.
It presents unique challenges due to the large viewpoint disparity and dynamic transformations inherent to V2D scenarios.
We again follow the official protocol and data splits for evaluation.
Consistent with our long-range focus, we expand the evaluation horizon from the original 50 meters to 100 meters.

\noindent\textbf{Evaluation Metrics.}
Both benchmarks adopt established metrics from the NuScenes benchmark~\cite{caesarNuScenesMultimodalDataset2020} and focus on the 'car' class.
We primarily report Average Precision (AP) to assess detection quality and Average Multi-Object Tracking Accuracy (AMOTA) for tracking performance.
To validate our claims of efficiency and practicality, we also measure transmission cost in Bytes Per Second (BPS) and computational efficiency via inference Frames Per Second (FPS).

\subsection{Implementation Details}
\label{subsec:implementation-details}

\noindent\textbf{Model and Baselines.} All models are built upon a ResNet50~\cite{heDeepResidualLearning2016} backbone, pre-trained on ImageNet.
For a fair comparison, baselines were reproduced as follows: UniV2X~\cite{yuEndtoEndAutonomousDriving2025}, CoopTrack~\cite{zhongCoopTrackExploringEndtoEnd2025}, and SparseCoop~\cite{wangSparseCoopCooperativePerception2026} were implemented using their official codebases.
Where2Comm~\cite{huWhere2commCommunicationefficientCollaborative2022} and V2X-VIT~\cite{xuV2XViTVehicletoeverythingCooperative2022} were re-implemented using a BEVFormer~\cite{liBEVFormerLearningBirdseyeview2022} backbone within the shared UniV2X/CoopTrack codebase.
For these BEV-based methods, we set the grid resolution to $1m \times 1m$, with the total grid size scaling proportionally to the perception range.
The query-based methods~\cite{yuEndtoEndAutonomousDriving2025,zhongCoopTrackExploringEndtoEnd2025,wangSparseCoopCooperativePerception2026} all use 900 queries.
Our Long-SCOPE framework adopts the same set, which is then augmented by our novel GQG module that generates at most 40 additional dynamic queries.
Our CAA module is implemented with 4 iterative refinement layers.

\noindent\textbf{Training Regimen.} 
All methods adopt the two-stage training strategy.
Initially, the single-agent models are trained for 48 epochs.
Subsequently, the cooperative model is initialized with the single-agent weights and fine-tuned for an additional 24 epochs.
We employ the AdamW optimizer with a total batch size of 16 and a weight decay of 0.01.
The initial learning rate is set to $2\times10^{-4}$ and is adjusted using a cosine annealing schedule.

\noindent\textbf{Data and Hardware.}
Input images from V2X-Seq are resized from $1920\times1080$ to $960\times540$,
while images from Griffin-25m are resized to $800\times448$.
All experiments were performed distributed across four NVIDIA 3090 GPUs.

\subsection{State-of-the-Art Comparison}
\label{subsec:sota-comparison}

\begin{table*}[t]
    \centering
    \caption{
    Detailed ablation study for the GQG module.
    We report the performance gains of single-agent models on \textit{Griffin-25m} dataset.
    }
    \label{tab:ablation-generation}
    \setlength{\tabcolsep}{5pt} 
    \begin{tabular}{clcccccc}
    \toprule
    \multirow{2}{*}[-0.5ex]{\centering\textbf{Agent Type}} & \multirow{2}{*}[-0.5ex]{\textbf{Query Method}} & \multicolumn{2}{c}{\textbf{0--100 m Overall}} & \multicolumn{2}{c}{\textbf{0--50 m}} & \multicolumn{2}{c}{\textbf{50--100 m}} \\
    \cmidrule(lr){3-4} \cmidrule(lr){5-6} \cmidrule(lr){7-8}
     &  & \textbf{AP} $\uparrow$ & \textbf{AMOTA} $\uparrow$ & \textbf{AP} $\uparrow$ & \textbf{AMOTA} $\uparrow$ & \textbf{AP} $\uparrow$ & \textbf{AMOTA} $\uparrow$ \\
    \midrule
    \multirow{3}{*}{\parbox{2.5cm}{\centering{Drone}\\(High-Vantage)}} & Baseline (Static Queries) & 0.371 & 0.393 & 0.639 & 0.766 & 0.092 & 0.010 \\
     & + Dynamic (Direct Depth) & 0.390 & 0.403 & 0.647 & 0.745 & 0.116 & 0.032 \\
     & + Dynamic (Height-Derived) & \rankone{0.449} & \rankone{0.443} & \rankone{0.694} & \rankone{0.748} & \rankone{0.182} & \rankone{0.099} \\
    \midrule
    \multirow{2}{*}{\parbox{2.5cm}{\centering{Vehicle}\\(Ground-Level)}} & Baseline (Static Queries) & 0.413 & 0.252 & 0.604 & 0.384 & 0.047 & \rankone{0.019} \\
     & + Dynamic (Direct Depth) & \rankone{0.428} & \rankone{0.256} & \rankone{0.610} & \rankone{0.407} & \rankone{0.052} & 0.000 \\
    \bottomrule
    \end{tabular}
\end{table*}

As detailed in~\cref{tab:griffin-25m,tab:v2x-seq}, Long-SCOPE achieves state-of-the-art performance on both the aerial-ground \textit{Griffin-25m} and real-world V2I \textit{V2X-Seq} datasets, significantly outperforming all baselines in both detection and tracking.
On \textit{Griffin-25m}, Long-SCOPE achieves an overall AP of 0.354, outperforming the next-best feature-level method (CoopTrack) by 7.5~AP points and the sparse baseline (SparseCoop) by 8.9~AP points.
Similarly, in the 0--150~m overall evaluation on \textit{V2X-Seq}, Long-SCOPE reaches 0.399~AP and 0.444~AMOTA, surpassing SparseCoop by 6.5~AP points and the dense V2X-ViT by 11.0~AP points.

Crucially, Long-SCOPE's advantage widens significantly as the perception range extends, proving its unique capability for true long-range fusion. 
In the challenging 50--100~m bracket on \textit{Griffin-25m}, Long-SCOPE maintains robust tracking (0.112~AMOTA) and achieves 0.151~AP, demonstrating graceful degradation.
In contrast, methods like CoopTrack and V2X-ViT collapse to near-zero tracking capabilities.
On \textit{V2X-Seq}, this resilience is even more pronounced:
in the extreme 100--150~m bracket, Long-SCOPE (0.113~AP) nearly doubles the performance of the closest competitor, CoopTrack (0.059~AP), and is 3.6 times more accurate than SparseCoop (0.031~AP). 
These results confirm our framework's effectiveness in resolving the severe positional noise and sparse observations inherent to extended distances.

\noindent\textbf{System Efficiency.}
Beyond accuracy, Long-SCOPE shows high practical viability for real-world deployment.
It maintains a highly competitive transmission cost of $1.90 \times 10^{5}$~BPS.
This is approximately 17 times more bandwidth-efficient than the dense BEV-based V2X-ViT ($3.21 \times 10^{6}$~BPS)
and remains on par with other sparse methods, delivering markedly superior perception without inflating communication overhead.
Furthermore, in terms of computational efficiency, Long-SCOPE achieves an inference speed of 7.68~FPS.
This is fully competitive with both dense and sparse baselines (e.g., V2X-ViT at 7.50~FPS and UniV2X at 6.51~FPS),
confirming that the substantial gains in long-range accuracy and robust query association do not come at the expense of real-time processing capabilities.

\subsection{Ablation Studies}
\label{subsec:ablation-studies}

To verify the individual contributions of our novel components, we conduct extensive ablation studies, with the results summarized in~\cref{tab:griffin-25m,tab:v2x-seq,tab:ablation-generation}.

\noindent\textbf{Impact of Core Components.}
We first validate the contributions of our two primary modules by removing them individually.
The Context-Aware Association (CAA) module proves to be the cornerstone for long-range fusion.
As shown in~\cref{tab:griffin-25m} and~\cref{tab:v2x-seq}, omitting the CAA module causes a severe performance drop, with the overall AP falling from 0.354 to 0.329 on \textit{Griffin-25m}, and from 0.399 to 0.352 on \textit{V2X-Seq}. 
Critically, this degradation is most severe precisely where Long-SCOPE claims its largest advantage: in the 100--150~m range, AP on \textit{V2X-Seq} plummets from 0.113 to 0.041.
This sharp decline, which reduces our model's performance to the level of competing baselines, provides direct evidence for our central thesis:
the simple, heuristic-based matching used by existing frameworks is the primary failure point under long-range alignment errors.
Our context-driven, learnable CAA module is the specific mechanism required to resolve this bottleneck.

Concurrently, the Geometry-guided Query Generation (GQG) module provides a foundational performance lift.
Removing it consistently degrades overall performance across both datasets (0.354 to 0.284~AP on \textit{Griffin-25m}, and 0.399 to 0.388~AP on \textit{V2X-Seq}).
This confirms that generating high-quality, data-driven queries is essential for initiating the perception pipeline with accurate proposals for distant targets, which subsequently enhances the entire multi-agent fusion process.

\begin{figure*}[ht]
    \centering
    \includegraphics[width=\textwidth]{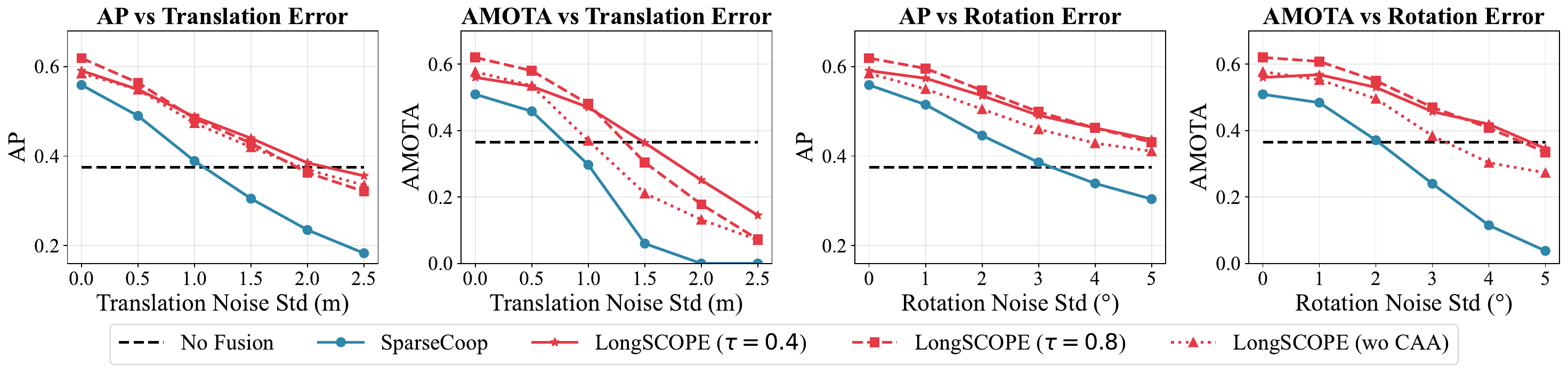}
    \caption{Comparison of robustness to localization errors on \textit{Griffin-25m} dataset. Long-SCOPE maintains stable tracking and detection under increasing translation and rotation noise compared to the baseline.}
    \label{fig:4_0_localization}
\end{figure*}

\begin{figure}[ht]
    \centering
    \includegraphics[width=\columnwidth]{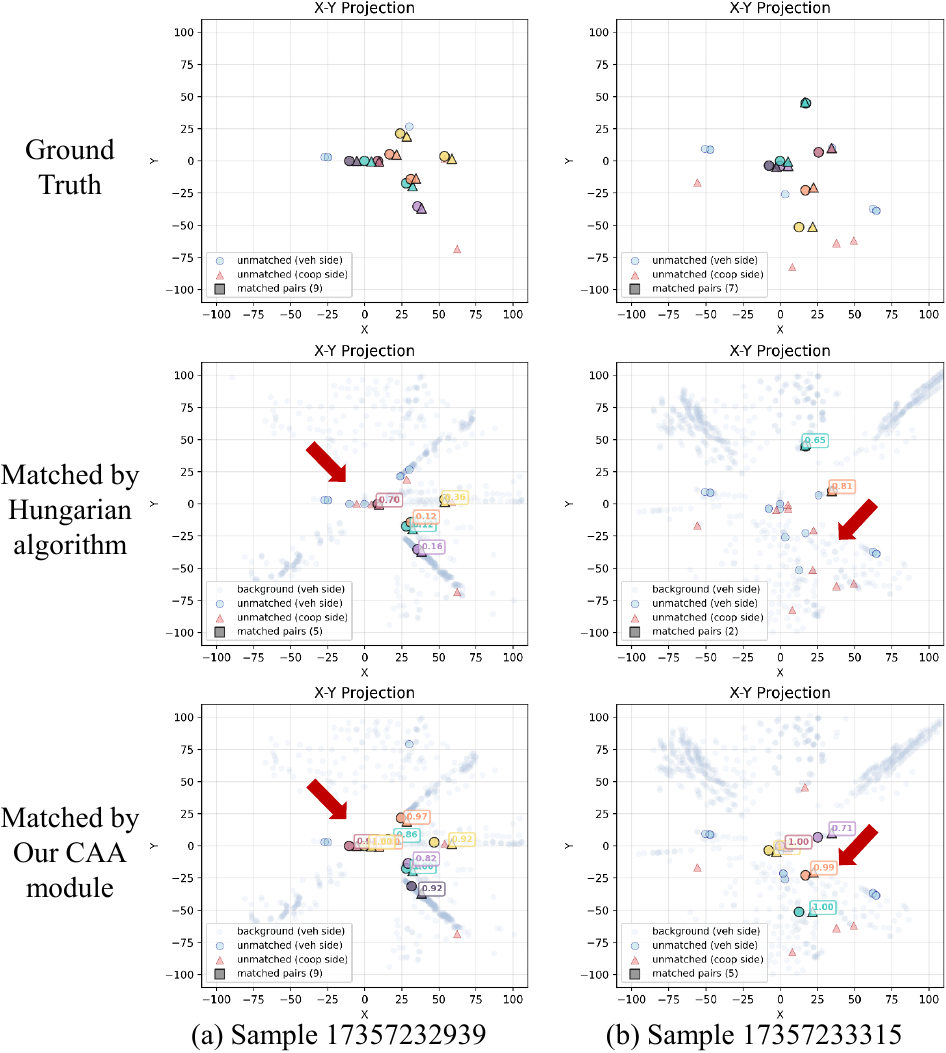}
    \caption{
        Qualitative comparison of query association performance under severe localization noise on \textit{Griffin-25m} dataset.
        Matched query pairs share the same color.
        While the Hungarian algorithm (middle, red arrows) struggles with alignment errors by prioritizing geometric closeness,
        our CAA module (bottom) robustly resolves these ambiguities.
    }
    \label{fig:4_1_CAA}
\end{figure}

\noindent\textbf{Validation of GQG Module Design.}
We further validate the GQG module in~\cref{tab:ablation-generation}.
For high-vantage agents (e.g., drones), our proposed height-derived depth estimation is vastly superior to direct regression.
It boosts the 50--100~m~AP from 0.116 (with Direct Depth) to 0.182, nearly doubling the long-range AP of the static baseline (0.092).
This validates the stability of predicting global height rather than depth. 
Conversely, for ground-level agents (e.g., vehicles) where this geometric derivation is unstable,
using standard direct depth regression for dynamic queries still provides a consistent, modest improvement over static queries alone (0.052 vs.\ 0.047~AP in the 50--100~m range).
This ablation confirms our two-pronged strategy within the GQG module is well-founded and optimal for diverse agent types.

\subsection{Robustness to Localization Noise}
\label{subsec:robustness}

Real-world V2X systems are inherently susceptible to GPS and calibration deviations.
To validate the practical resilience of the CAA module against these alignment errors, we investigate Long-SCOPE's performance under increasing levels of injected localization noise in~\cref{fig:4_0_localization}.
We observe a distinct trade-off regarding the matching confidence threshold $\tau$.
While a stricter threshold ($\tau=0.8$) provides peak performance under ideal, noise-free localization, its criteria become overly conservative as noise increases, inadvertently filtering out some valid cooperative queries.
Conversely, a more balanced threshold ($\tau=0.4$) demonstrates superior real-world robustness, maintaining reliable association capabilities even under significant alignment error, including translation noise with a standard deviation exceeding 1.0~m or rotation noise beyond $4^{\circ}$.
Furthermore, Long-SCOPE equipped with the CAA module clearly outperforms both the baseline variant using standard Hungarian matching (denoted as \textit{w/o CAA}) and the original SparseCoop~\cite{wangSparseCoopCooperativePerception2026} architecture.

\Cref{fig:4_1_CAA} visualizes the critical failure modes of standard association techniques compared to our proposed CAA module.
Standard bipartite matching (e.g., the Hungarian algorithm~\cite{kuhnHungarianMethodAssignment1955}) relies almost exclusively on Euclidean distance. 
As illustrated in the middle row of~\cref{fig:4_1_CAA}, when significant localization noise is introduced, the projected position of a cooperative agent's query drifts far from the ego query.
Consequently, the Hungarian algorithm forces matches based on geometric proximity, often incorrectly associating distinct objects or completely failing to match corresponding ones.
The CAA module overcomes this limitation by aggregating local neighborhood information via self-attention prior to the matching step. 
Considering that a cluster of vehicles maintains its relative shape and semantic signature even when absolute coordinates are globally shifted,
this preservation of structural integrity allows Long-SCOPE to correctly associate queries despite severe coordinate misalignment.
\section{Conclusion}
\label{sec:conclusion}

We introduced Long-SCOPE, a fully sparse framework for long-range cooperative 3D perception that avoids the quadratic scaling costs of BEV-based methods.
Our approach features two novel components: Geometry-guided Query Generation to accurately detect distant targets, and a learnable Context-Aware Association module that replaces fragile, heuristic-based matching.
Experiments on \textit{V2X-Seq} and \textit{Griffin-25m} show Long-SCOPE achieves state-of-the-art performance, with its advantage widening dramatically at extreme distances (100--150~m) where existing methods collapse.
This work validates that a fully sparse design, centered on robust, context-aware association, is essential for practical long-range cooperative perception.

\section*{Acknowledgments}
This work was supported by the National Natural Science Foundation of China for the Science Fund for Creative Research Groups (No. 52221005), the Key Project (No. 52131201), and the Fundamental and Interdisciplinary Disciplines Breakthrough Plan of the Ministry of Education of China (No. JYB2025XDXM123).

{
    \small
    \bibliographystyle{ieeenat_fullname}
    \bibliography{THICV}
}

\end{document}